\documentclass[conference]{IEEEtran}
\IEEEoverridecommandlockouts

\usepackage{bm}
\usepackage{xcolor}
\usepackage{amsmath}
\usepackage{amssymb}
\usepackage{siunitx}
\usepackage{graphicx}
\usepackage{hyperref}
\usepackage{mathtools}

\usepackage{tikz}
\usepackage{pgfplots}
\pgfplotsset{compat=newest}
\usetikzlibrary{patterns}

\def\equationautorefname~#1\null{Eq.~(#1)\null}
\def\chapterautorefname~#1\null{Chapter~#1\null}%
\def\sectionautorefname~#1\null{Sec.~#1\null}%
\def\subsectionautorefname~#1\null{Sec.~#1\null}%
\def\figureautorefname~#1\null{Fig.~#1\null}%
\def\tableautorefname~#1\null{Table~#1\null}%

\newcommand{\longautoref}[1]{%
	\begingroup%
	\def\equationautorefname~##1\null{Equation~(##1)\null}%
	\def\chapterautorefname~##1\null{Chapter~##1\null}%
	\def\sectionautorefname~##1\null{Section~##1\null}%
	\def\subsectionautorefname~##1\null{Subsection~##1\null}%
	\def\figureautorefname~##1\null{Figure~##1\null}%
	\def\tableautorefname~##1\null{Table~##1\null}%
	\autoref{#1}%
	\endgroup%
}

\ifCLASSOPTIONcompsoc
\usepackage[caption=false,font=normalsize,labelfont=sf,textfont=sf]{subfig}
\else
\usepackage[caption=false,font=footnotesize]{subfig}
\fi

\usepackage{algorithm}
\usepackage{algorithmicx}
\usepackage{algpseudocode}
\usepackage[outline]{contour}
\contourlength{1.75pt}
\usepackage{float}

\newcommand\copyrighttext{\footnotesize \textcopyright~2023 IEEE. Personal use of this material is permitted. Permission from IEEE must be obtained for all other uses, in any current or future media, including reprinting/republishing this material for advertising or promotional purposes, creating new collective works, for resale or redistribution to servers or lists, or reuse of any copyrighted component of this work in other works. DOI: \href{https://ieeexplore.ieee.org/document/10186535}{10.1109/IV55152.2023.10186535}
}

\newcommand\copyrightnotice{%
    \begin{tikzpicture}[remember picture,overlay]%
     \node[anchor=south, xshift=0pt, yshift=20pt] at (current page.south)%
     {\fbox{\parbox{\dimexpr\textwidth-\fboxsep-\fboxrule\relax}{\copyrighttext}}};%
     \end{tikzpicture}%
}

\def\BibTeX{{\rm B\kern-.05em{\sc i\kern-.025em b}\kern-.08em
    T\kern-.1667em\lower.7ex\hbox{E}\kern-.125emX}}
    
\begin{document}

\title{Real-Time Spatial Trajectory Planning for Urban Environments Using Dynamic Optimization}

\author{\IEEEauthorblockN{Jona Ruof, Max Bastian Mertens, Michael Buchholz, and Klaus Dietmayer}%
\IEEEauthorblockN{
\textit{Institute of Measurement, Control, and Microtechnology, Ulm University}, Germany \\
\tt{\{firstname\}.\{lastname\}@uni-ulm.de}}%
\thanks{
Part of this work has been financially supported by the Federal Ministry for Economic Affairs and
Climate Action of Germany within the program ”Highly and Fully Automated Driving in Demanding
Driving Situations” (project LUKAS, grant number 19A20004F).}
}

\maketitle

\begin{abstract}
Planning trajectories for automated vehicles in urban environments requires
methods with high generality, long planning horizons, and fast update rates.
Using a path-velocity decomposition, we contribute a novel planning
framework, which generates foresighted trajectories and can handle a wide
variety of state and control constraints effectively. In contrast to
related work, the proposed optimal control problems are formulated over space
rather than time. This spatial formulation decouples environmental constraints
from the optimization variables, which allows the application of simple, yet
efficient shooting methods. To this end, we present a tailored solution
strategy based on ILQR, in the Augmented Lagrangian framework, to rapidly
minimize the trajectory objective costs, even under infeasible initial solutions.
Evaluations in simulation and on a full-sized automated vehicle in real-world
urban traffic show the real-time capability and versatility of the proposed approach.
\end{abstract}

\begin{IEEEkeywords}
trajectory planning, automated driving, augmented lagrangian, iterative lqr
\end{IEEEkeywords}

\section{Introduction}

\copyrightnotice

Trajectory planning forms the foundation for numerous applications in automated
driving. Recently, the development has been increasingly focused on planning
in urban environments \cite{Artunedo2019,Graf2022,Micheli2022}.
This has proven challenging, as the interdependence between traffic participants and the
resulting problem complexity often make situations hard to predict. In order to still
react safely in critical situations, trajectory updates need to be computed with high
frequency. Additionally, the situational diversity
of urban traffic imposes varied space, velocity, acceleration, or even spatiotemporal
constraints, making methods with high generality necessary. Moreover, the comfort and safety
of the resulting trajectory are often directly dependent on the considered horizon. 
Computing foresighted trajectories thereby requires planning over intervals $>\SI{10}{s}$
or often up to the perception range of the vehicle ($> \SI{100}{m}$). The
requirement of generality combined with long-horizon planning contradicts fast reaction
times, thus constituting a conflict that needs to be addressed.

Towards the outlined dilemma, we investigate the formulation of a trajectory planning
framework, which is designed to run at high frequencies (e.g. \SI{100}{Hz}) and
is versatile enough to handle driving in urban environments. In contrast to most
related work, we explore the optimization of trajectories over \emph{space} rather
than \emph{time}. This choice is motivated by the observation that in automated driving,
many situations (e.g. speed limits, stop signs, traffic lights, ...) impose constraints
expressed as functions of space $s$. By defining the trajectory likewise as
a function of space, the constraint value is decoupled from the resulting state
$\bm{x}(s)$ and control variables $\bm{u}(s)$ in the trajectory. This is in
contrast to a temporal formulation, where, e.g., $s(t=\SI{10}{s})$ at the end of the
trajectory is dependent on the choice of the acceleration $a(t=\SI{0}{s})$ at the
beginning of the trajectory. For locally fixed constraints, the spatial formulation
stabilizes the optimization, even permitting the use of efficient single-shooting
methods (such as AL-ILQR), which are often considered less practical for longer time
horizons \cite{Giftthaler2018}.

Structurally, we propose to apply a path-velocity decomposition approach
(\autoref{fig:sim-scenario}), where this work is focused on longitudinal optimization.
Our main contributions are:

\begin{figure}
    \includegraphics[width=\columnwidth]{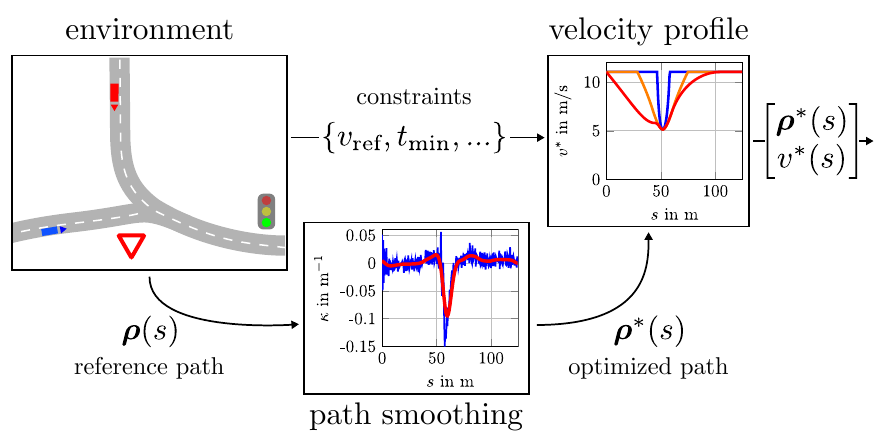}
    \vspace{-0.7cm}
    \caption{Overview of the proposed approach given an exemplary urban scenario.
             The ego vehicle (blue) has to reach a traffic light in time while respecting
             the right of way of the other vehicle (red). We propose a spatial, decomposed
             path-velocity formulation (\autoref{sec:objectives}) which is solved with a dynamic optimization approach (\autoref{sec:solution_strategy}).}
    \label{fig:sim-scenario}
\end{figure}

\begin{itemize}

\item A dynamic optimization formulation, which can be solved in
the millisecond range ($<\SI{10}{ms}$) and generates foresighted
trajectories over a \SI{125}{m} horizon. Additionally, static and dynamic obstacles,
velocity, acceleration, curvature limits, as well as spatiotemporal constraints, are
considered.

\item A tailored, real-time capable solution strategy based on ILQR in the
Augmented Lagrangian framework, with improved optimization stability and
implicit treatment of infeasible initializations.

\item An evaluation in simulation and on a full-sized automated
vehicle in real urban traffic, demonstrating the versatility and
robustness of the presented approach.

\item An open-source implementation in C and Python along with simulated experiments,
which can be found at \href{https://github.com/joruof/tpl}{https://github.com/joruof/tpl}.

\end{itemize}

The remainder of this paper is structured as follows: after discussing related
work in \autoref{sec:related_work}, the two central dynamic optimization
objectives are presented and motivated in \autoref{sec:objectives}. Then the solution
strategy is outlined in \autoref{sec:solution_strategy}, and the presented approach
is evaluated in simulation and real-world experiments in \autoref{sec:evaluation}.
We conclude with a brief discussion of current limitations and directions for
future work in \autoref{sec:conclusion}.

\section{Related Work}
\label{sec:related_work}

Literature related to our contribution can be grouped into three main categories,
which are discussed in this section.

\subsection{Urban Trajectory Planning Frameworks}

While many trajectory planning strategies for vehicles have been developed,
most publications focus on specific scenarios and only few propose general methods
for urban environments. End-to-end learning methods such as
\cite{Casas2021} show promising results, but often consider short horizons and
exhibit generalization issues. Recently, \cite{Artunedo2019} introduced
a general motion planning framework based on B\'{e}zier sampling, which was proven
effective in real-world scenarios. Ziegler et al. \cite{Ziegler2014} proposed
local-continuous optimization to obtain trajectories in a general planning
framework, which is able to drive a car on public urban roads. For
long horizons, graph search methods in combination with continuous
optimization have been demonstrated by Hubmann et al. \cite{Hubmann2016} for
longitudinal planning. Our previous work \cite{Graf2022}, similarly to
\cite{Ziegler2014}, also relied on continuous optimization supported by driver
models and performed well on public urban roads. Another recent work is
\cite{Micheli2022}, using local-continuous optimization in an MPC (Model Predictive Control)
fashion and showing promising results for overtaking and cut-in scenarios in simulation.
In contrast to our work, all listed publications either update the trajectory
with lower frequency (\cite{Hubmann2016,Graf2022}\,@\,\SI{10}{Hz};
\cite{Micheli2022,Artunedo2019}\,@\,\SI{20}{Hz}; \cite{Ziegler2014} terminates after 
\SI{0.5}{s}), consider only a short horizon (max. \SI{3}{s} in \cite{Micheli2022}),
or do not handle explicit spatiotemporal constraints.

\subsection{Iterative LQR}

The iterative LQR (ILQR) algorithm \cite{Tassa2012} is a single-shooting,
discrete-time, local-continuous trajectory optimization method. It has been
successfully applied to optimal control in various robotic applications, though
usually with short horizons for e.g. trajectory tracking \cite{Nagariya2020,Chen2018B}.

Further ILQR variants have been proposed to handle, e.g., multiple-shooting objectives
\cite{Giftthaler2018} or control constraints \cite{Tassa2014}. State constraints have been,
e.g., handled in an interior point variant \cite{Pavlov2021} or in a simpler approach
called Constrained Iterative LQR (CILQR), proposed in \cite{Chen2018A} and further
analyzed in \cite{Chen2019}. CILQR uses barrier functions to enforce state
constraints and has already been successfully applied to automated driving:
While initially being developed for obstacle avoidance in \cite{Chen2018A}, it
has been extended with stochastic prediction and weight-tuning to handle more
diverse situations, such as cut-ins \cite{Chen2019}. Moreover,
\cite{Shimizu2021} improves numerical issues in CILQR and introduces initial
trajectory guesses, while \cite{Jahanmahin2021} utilizes CILQR for planning
with reduced jerk and also explores automatic weight tuning.

While the barrier functions used in CILQR are comparatively simple to
understand and implement, they suffer from numerical issues in case of
infeasible initialization (cf.~\cite{Shimizu2021}) and must be tuned
either manually or with an automatic weight-tuning scheme (cf.
\cite{Jahanmahin2021,Chen2019}). Recently, these issues have been
addressed by incorporating ILQR in an ADMM (Alternating Direction Method of Multipliers)
method \cite{Ma2022}. Another numerically robust, but simpler alternative for enforcing
constraints is the Augmented Lagrangian (AL) method, as implemented by the ALTRO package
\cite{Howell2019} as AL-ILQR. Although AL-ILQR strikes a balance between
implementation/runtime complexity and constraint satisfaction, it has
not yet been adapted for long-horizon trajectory planning in automated driving.

\subsection{Spatial Trajectory Planning}

The development of long-horizon, spatial trajectory planning is mainly
driven by research in the domain of automated trucking. Earlier works such as
\cite{Hellstroem2009} already proposed spatial trajectories to compute
long-horizon trajectories for reduced fuel consumption and were recently extended
in \cite{Bae2019}. In contrast to our work, these works obtain 
trajectories through dynamic programming and also update with much lower rates
(e.g., \SI{200}{ms} in \cite{Bae2019}). Another recent work \cite{Voswinkel2020} similarly
applies a path-velocity decomposition and plans trajectories in the spatial domain.
In contrast to our online approach, they solve part of the planning problem offline
and do not consider spatiotemporal constraints.

\section{Formulation}
\label{sec:objectives}

The following section outlines general path and velocity objectives
defined over \emph{space} and shows how constraints are represented
in the objectives.

\subsection{Path Objective}
\label{sec:path-opt}

We assume access to a reference path $\bm{\rho}(s) = [\hat{x}(s), \hat{y}(s)]^T$
given as a vector-valued function in a 2D Cartesian frame over space $s$ from,
e.g., a prerecorded GPS map or a lane detection module.
As the path curvature $\kappa(s)$ directly limits the maximum velocity, accurate
estimation is crucial for the generation of a smooth velocity profile. Because local
curvature computation yields noisy results \cite{Gritschneder2018}, we adopt the
formulation from \cite{Gritschneder2018}.
Thereby, a path with smoothed curvature is found as a solution to an optimal control
problem with state $\bm{x}_r = [x_r, y_r, \phi_r]^T$ and control
$u_r = \kappa$. The objective dynamics $\partial\bm{x}_r / \partial s = 
\bm{f}_r(\bm{x}_r, u_r)$ are given by
\begin{align}
\label{eq:path_cost_function}
    \frac{\partial x_r}{\partial s} &= \cos(\phi_r)\,, &
    \frac{\partial y_r}{\partial s} &= \sin(\phi_r)\,, &
    \frac{\partial \phi_r}{\partial s} &= \kappa\,,
\end{align}
which are integrated from the boundary condition $\bm{x}_r(0) =
\bm{x}_{r,0}$ over space $s \in [0, S]$, with $S$ being the path horizon.
Trajectories obtained through the integration are rated by the quadratic function (dropping
$s$ for sake of readability) 
\begin{align}
    l_r(\bm{x}_r, u_r) 
    &= w_d \cdot ||(\hat{x} - x_r, \hat{y} - y_r)||_2^2
    + w_\kappa \cdot ||\kappa||_2^2\,, 
\end{align}
with $w_d, w_\kappa$ being weights on the error of the Cartesian coordinates
and the applied curvature, respectively. Including box-constraints $\kappa \in
[\kappa_\textrm{min}, \kappa_\textrm{max}]$ to ensure dynamic feasibility, the
objective formulates as
\begin{subequations}
\label{eq:ref_line_objective}
\begin{align}
    \min_{\kappa(\cdot)} \int_0^S & l_r(\bm{x}_r, u_r)\,\textrm{d}s \quad \textrm{s.t.}
    & \frac{\partial\bm{x}_r}{\partial s_r} = \bm{f}_r(\bm{x}_r, u_r)\,, \\
    & \bm{x}_r(0) = \bm{x}_{r,0}\,,
    & \kappa_\textrm{min} \leq \kappa \leq \kappa_\textrm{max} \, .
\end{align}
\end{subequations}
For a minimizer $\kappa^*(s)$ of the objective, a driveable reference path
with smoothed curvature $\bm{\rho}^*(s)$ is then obtained by the respective integrals of $x_r$, $y_r$ over $s$.

\subsection{Velocity Objective}
\label{sec:velocity-objective}

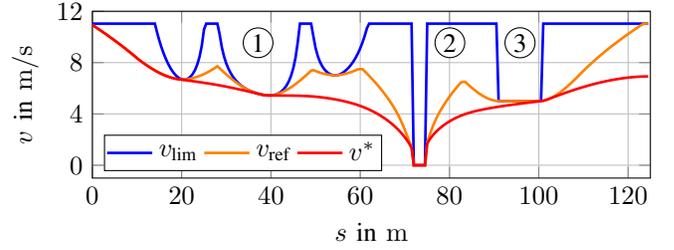
\begin{figure}[tbp]
\centering
\begin{tikzpicture}
\begin{axis}[
    title={},
    width=9cm,
    height=3.8cm,
    xlabel={$s$ in \SI{}{m}},
    ylabel={$v$ in \SI{}{m/s}},    
    xlabel near ticks,
    ylabel near ticks,
    xmin=0, xmax=125,
    ymin=-1, ymax=12,
    xtick={0,20,40,60,80,100,120},
    ytick={0,4,8,12,16},
    xmajorgrids=true,
    ymajorgrids=true,
    grid style=solid,
    legend columns=-1,
    legend style={at={(0.27, 0.26)},
        anchor=north,
        inner sep=0.5pt,
        style={column sep=0.0cm}
    }
]
\addplot[
    color=blue,
    line width=1.0pt
    ]
    table[x=space,y=lim,col sep=comma] {data/example_vp.csv};
\addlegendentry{$v_\textrm{lim}$} 
\addplot[
    color=orange,
    line width=1.0pt
    ]
    table[x=space,y=ref,col sep=comma] {data/example_vp.csv};
\addlegendentry{$v_\textrm{ref}$} 
\addplot[
    color=red,
    line width=1.0pt
    ]
    table[x=space,y=opt,col sep=comma] {data/example_vp.csv};
\addlegendentry{$v^*$}
\end{axis}
\node[circle, draw=black, fill=white, anchor=west, inner sep=1pt, outer sep=3pt] at (1.9, 1.8) {1};
\node[circle, draw=black, fill=white, anchor=west, inner sep=1pt, outer sep=3pt] at (4.45, 1.8) {2};
\node[circle, draw=black, fill=white, anchor=west, inner sep=1pt, outer sep=3pt] at (5.38, 1.8) {3};
\end{tikzpicture}
\vspace{-0.7cm}
\caption{Intermediate steps of the velocity profile generation:
Based on $v_\textrm{lim}$, an acceleration and jerk limited profile $v_\textrm{ref}$
is generated, then optimized to yield  a foresighted $v^*$ profile.
Marker 1 shows an exemplary velocity constraint through road curvature.
Markers 2 and 3 show interaction constraints.
}
\label{fig:vel-prof-example}
\end{figure}

Starting from $a(t) = \partial v/ \partial t$, the application of the chain rule allows the
definition
of velocity over space dynamics
\begin{align}
    \label{eq:vel_over_space_dynamics}
    a(t) &= \frac{\partial v}{\partial t} = \frac{\partial v}{\partial s}
           \frac{\partial s}{\partial t} = \frac{\partial v}{\partial s} v(t)
    &&
    \Leftrightarrow
    &
    \frac{\partial v}{\partial s} &= \frac{a(t)}{v(t)} \,\,\,.
\end{align}
Analogously, spatial dynamics for jerk $j$ and time $t$ are 
$\partial j / \partial s = j(t) / v(t)$, $\partial t / \partial s = 1 / v(t)$.
One obvious problem with this dynamics function is the singularity at $v=0$.
Numerical issues are prevented by limiting $v \geq v_\textrm{min}$,
where $v_\textrm{min}$ is some small positive value. Velocities at $v_\textrm{min}$
are set to $0$ in a postprocessing step, if the target velocity $v_\textrm{lim} = 0$.
In general form, we assume that target/maximum velocities are specified over space as
$v_\textrm{lim}(s)$. The generation of $v_\textrm{lim}(s)$ depending on the concrete
traffic situation is detailed in the following subsections. From $v_\textrm{lim}$
we generate a reference profile $v_\textrm{ref}$ and a velocity profile $v^*$,
as depicted in Fig. \ref{fig:vel-prof-example}. This profile generation process is explained now.

As $v_\textrm{lim}(s)$ may not respect jerk or acceleration
constraints $[j_\textrm{min}, j_\textrm{max}]$, $[a_\textrm{min},
a_\textrm{max}]$, the constraint profile first is shaped by integrating
\begin{align}
    \label{eq:backward-forward}
    \left( \frac{\partial v}{\partial s}, \frac{\partial a}{\partial s} \right)^T &= 
    \begin{cases}
        \left(a/v, \, j/v \right)^T &: v \leq v_\textrm{lim} \\
        (0, 0)^T &: \textrm{else} \\ 
    \end{cases} 
\end{align}
once backwards from horizon $S$ to $0$, with $j(s) = j_\textrm{min}$, boundary
condition $(v(S), a(S))^T = (v_\textrm{lim}(S), 0)$, and once forwards
with $j(s) = j_\textrm{max}$, boundary condition $(v(0), a(0))^T =
(v(0), 0)$. Additionally, during integration of the resulting
profiles $v_\textrm{bwd}$ and $v_\textrm{fwd}$, states $v, a$ are limited after
every integration step to $v = \min(v, v_\textrm{lim})$ and $a =
\textrm{clip}(a_\textrm{min}, a_\textrm{max}, a)$.
Application of the minimum operator then yields a profile
\begin{align}
    v_\textrm{ref}(s) &= \min(v_\textrm{bwd}(s), v_\textrm{fwd}(s))\,,
\end{align}
which respects acceleration and jerk limits, but is neither foresighted nor comfortable.
Therefore, $v_\textrm{ref}$ is subsequently incorporated into a dynamic
optimization objective as a velocity target and as an upper limiting function for feasible profiles. 

To this end, we define $\bm{x}_v = [v, t]^T$,
$u_v = a$, and $f_v(\bm{x}_v, u_v) = [\partial v /
\partial s, \partial t / \partial s]^T$ with
boundary condition $v(0) = v_\textrm{veh}$. Profiles $v(s)$ generated by
forward integration of $f_v(\bm{x}_v, u_v)$ are rated by the
quadratic function
\begin{align}
    \label{eq:profile_rating}
    l_v(\bm{x}_v, u_v) &=
    w_v ||v(s) - v_\textrm{ref}(s)||^2_2 + w_a ||a(s)||^2_2
\end{align}
with weights $w_v, w_a$. Combining dynamics $f_v$ and \eqref{eq:profile_rating}
yields a dynamic minimization objective over velocity profiles
\begin{subequations}
\label{eq:profile_objective}
\begin{align}
    \min_{a(\cdot)} \, \int_0^S & l_v(\bm{x}_v, u_v) \, \textrm{d}s \quad 
        \textrm{s.t.} \quad & \frac{\partial \bm{x}_v}{\partial s} = f_v(\bm{x}_v, u_v)\,, \\
    & v_\textrm{min} \leq v \leq v_\textrm{ref}\,, 
    & a_\textrm{min} \leq a \leq a_\textrm{max} \label{eq:profile_constr}\,,
\end{align} 
\end{subequations}
where the optimal profile $v^*(s)$ is obtained by the integral of a minimizer $a^*$ of this objective using the dynamics
\eqref{eq:vel_over_space_dynamics}.

While the cost function $l_v$ takes a simple quadratic form, the objective
becomes intricate through its non-linear dynamics and the presence of
control and state constraints \eqref{eq:profile_constr}. Thus, a specialized algorithm is required
to solve the objective efficiently, which we propose in \autoref{sec:solution_strategy}.

\subsection{Speed Limits}

In the case of urban automated driving, the velocity limit function $v_\textrm{lim}(s)$
(\autoref{fig:vel-prof-example})
is shaped by legal speed limits $v_\textrm{lg}$ and implicit velocity limits
induced by the curvature $\kappa^*(s)$ of the road. Given a maximum admissible lateral
acceleration $\hat{a}_\textrm{lat}$, $v_\textrm{lim}$ is initially given by 
\begin{align}
v_\textrm{lim}(s) &= \min \left(v_\textrm{lg}(s), \sqrt{|\hat{a}_\textrm{lat}|/|\kappa^*(s)|} \right) \, ,
\end{align}
where $\kappa^*(s)$ is obtained as solution of objective \eqref{eq:ref_line_objective}.
\subsection{Interaction Constraints}
Static interaction constraints, which are fixed to a certain location $s_o$, 
are expressed simply by updating $v_\textrm{lim}$ with the minimum operator
\begin{align}
    v_\textrm{lim}(s_o) &\leftarrow \min(v_\textrm{lim}(s_o), v_o) \,,
\end{align}
where $v_o = 0$ for interactions that require stopping (e.g. traffic lights) and
$0 \leq v_o \leq v_\textrm{lim}$ for interactions with velocity reductions.
Note that through the backward-forward integration presented in
\eqref{eq:backward-forward}, even significant changes in $v_\textrm{lim}$ (e.g. through
obstacles) still result in a smooth reference profile $v_\textrm{ref}$ (\autoref{fig:vel-prof-example}).
Similarly to static interactions, we consider generic dynamic interactions (e.g., cars,
bicycles, ...) at position $s_o$ with velocity $v_o$ as constraints in the velocity profile.
The obstacles are expressed by a reactive velocity constraint formulation, motivated by a
constant velocity prediction. Given a safety distance $d_\textrm{safe}$ (which may
depend on $v$), the constraint updates $v_\textrm{lim}$ according to
\begin{align}
    v_\textrm{lim}(s) &\leftarrow 
        \min(v_\textrm{lim}(s), v_o \cdot \min(1, s_o / d_\textrm{safe}))  \, ,
\end{align}
at all $s \in [s_o - d_\textrm{safe}, s_o]$. The resulting velocity
constraint ensures that $v = v_o$ at the beginning of the safety distance and
then reduces $v$ proportional to the amount of safety distance violation
until standstill at $s = s_o$.
This heuristics ensures that in case of a decelerating obstacle,
the distance is increased and approaches the safety distance, provided
the obstacle decelerates no more than $a_\textrm{min}$.
Even in case of a slow obstacle ahead, a collision is avoided if there is any
viable solution given the same dynamics limits.
While this way of handling obstacles may seem simplistic, it results
in safe and smooth behavior, assuming fast replanning cycles.

\subsection{Spatiotemporal Constraints}

Spatiotemporal constraints (like in \autoref{fig:sim-scenario}) restrict
the ego vehicle to a certain time interval at a certain location.
Maximum spatiotemporal constraints require that the vehicle reaches a point
$s$ at time $t(s) \leq t_\textrm{max}(s)$, which may increase acceleration and
velocity. This is always feasible if the velocity stays below
$v_\textrm{ref}$ and the acceleration within $[a_\textrm{min}, a_\textrm{max}]$. Thus,
$t_\textrm{max}(s)$ constraints can be included simply as additional state
constraints in objective \eqref{eq:profile_objective}.
In contrast, minimum spatiotemporal constraints require that $t(s) \geq
t_\textrm{min}(s)$, which may reduce velocity until standstill.
Naively applying $t_\textrm{min}$ constraints without additional
regularization often causes stopping far ahead of the constraint location.
While this solution is valid, it is often perceived as unintuitive or
obstructive by other traffic participants. Therefore, we propose to express
a $t_\textrm{min}$ constraint at $s_c$ via the inequality
\begin{align}
    \label{eq:t_min_constr}
    (t_\textrm{min}(s_c) - t(s_c)) \cdot (v(s_c) - v_\textrm{min}) \leq 0\,.
\end{align}

The left side contains the constraint as-is in the first factor but also
approaches zero by reducing the velocity to $v_\textrm{min}$ in the second factor.
Inequality \eqref{eq:t_min_constr} therefore expresses, in a continuous way, that
both reducing the time at $s_c$ to $t_\textrm{min}$ as well as stopping at $s_c$
lead to constraint satisfaction. Additionally, we replace the weight on the
reference velocity deviation $w_v$ in \eqref{eq:profile_rating} with $w_v(s)$ shaped by
\begin{equation}
    \label{eq:t_min_weight_func}
    w_v(s) = \min(1, ((s - s_c - \alpha) \cdot \beta)^2 )\,.
\end{equation}
For a given offset $\alpha \geq 0$ and scale $0 < \beta < 1$, this weighting
function mitigates the conflict between driving at $v_\textrm{ref}$ and
reducing $v$, so that $t_\textrm{min}(s_c)$ holds. As the function is minimal at or
above $s_c$, it encourages stopping precisely at $s_c$. In our experiments,
including both \eqref{eq:t_min_constr} and \eqref{eq:t_min_weight_func} in 
objective \eqref{eq:profile_objective} leads to decisive constraint handling
behavior, either slowing down before the constraint or stopping precisely at $s_c$.

\section{Solution Strategy}
\label{sec:solution_strategy}

\begin{algorithm}[!t]
\caption{Solution Strategy}
\label{alg:solution-strategy}
\begin{algorithmic}
\State initialize $n \leftarrow 1$ and $\mu_i$ with fixed large values
\State initialize $\bm{X}, \bm{U}, \bm{\lambda}$ shifted from previous step
\While{$n < N \, \textbf{and} \, \textrm{cost change} > \textrm{tol}$}
    \State minimize objective \eqref{eq:al_ilqr} with ILQR iteration
\EndWhile
\State $\lambda_{k,i} \leftarrow \lambda_{k,i} + \mu_{i} h_{i}(\bm{x}_k, \bm{u}_k)$
\State $\lambda_{k,i} \leftarrow \max(0, \min(\lambda_{k,i}, \lambda_{\textrm{max}, i}))$
\end{algorithmic}
\end{algorithm}

The solution strategy is based on ILQR, which is a single-shooting,
discrete-time, local-continuous trajectory optimization method for
unconstrained dynamic optimization objectives with general non-linear dynamics
and cost functions (cf.~\cite{Tassa2012}).
We argue that ILQR is a suitable choice for automotive applications as it, firstly, 
can be implemented efficiently for small control spaces in plain C code
(e.g. on embedded devices) even without specialized numeric routines. Secondly,
ILQR runtime complexity scales only linearly with the optimization horizon,
which makes it suitable for long-horizon planning. Thirdly, being a shooting method,
trajectories generated with ILQR are by design always dynamically consistent,
even when the optimization is terminated early. This allows one to trade off
reaction time for a temporarily suboptimal solution, which is subsequently
improved over multiple iterations (for convergence results cf.~\cite{Graichen2010}).

\subsection{Augmented Lagrangian ILQR}

As demonstrated by \cite{Howell2019}, ILQR can be embedded in an Augmented
Lagrangian method to solve discretized, constrained dynamic optimization
objectives
\begin{align}
    \label{eq:al_ilqr}
    \min_{\bm{u}_1, ..., \bm{u}_K} & \mathcal{L}(\bm{X},
        \bm{U}, \bm{\lambda}, \bm{\mu})  \quad \textrm{s.t} \quad \bm{x}_{k+1} = f(\bm{x}_k, \bm{u}_k),
        \\ & h_i(\bm{x}_k, \bm{u}_k) \leq 0 \nonumber \, , i \in [1, ..., I], \, \forall k \in [1, ..., K]\,,
\end{align}
where $\bm{X} = (\bm{x}_1, ..., \bm{x}_K)$, $\bm{U} =
(\bm{u}_1, ..., \bm{u}_K)$ are matrices of states and controls,
$\bm{\lambda} \in \mathbb{R}^{K \times I}$ is a matrix of
Lagrange multiplier estimates, $\bm{\mu} \in \mathbb{R}^I$ is a vector of barrier weights,
$f$ is the dynamics function, $h_i$ are $I$ constraint functions, and $k$ is
the summation index up to a horizon $K$. Further, $\mathcal{L}$ denotes the Lagrangian
\begin{align}
    \mathcal{L}(& \bm{X}, \bm{U}, \bm{\lambda}, \bm{\mu}) = \\ &\sum_{k=1}^K \, \left[ l(\bm{x}_k, \bm{u}_k)
            + \sum_{i=1}^I \left[ \lambda_{k,i} h_i(\bm{x}_k, \bm{u}_k) 
            + \mu_i h_i(\bm{x}_k, \bm{u}_k)^2 \right]  \right] \nonumber\,,
\end{align}
where $l$ is an integral cost. The Augmented Lagrangian ILQR (AL-ILQR) then
alternates between solving objective \eqref{eq:al_ilqr} up to a given tolerance
with standard ILQR, increasing the Lagrange multipliers by the amount of
weighted constraint violation, and monotonically increasing the quadratic barrier
weights until the constraints hold within tolerances (cf.~\cite{Howell2019}).

\subsection{AL-ILQR for Automotive Trajectory Planning}

The AL-ILQR approach in \cite{Howell2019} is focused on robotic applications,
where keeping constraint violations small is crucial. When adapting AL-ILQR
for automotive trajectory planning, we argue that rather reaction time is
critical, while small numerical temporary constraint violations are tolerable.
Algorithm~\autoref{alg:solution-strategy} lists the adapted solution strategy.
Because Alg.~\autoref{alg:solution-strategy} is executed at high rate, we found that
a single $\bm{\lambda}$ update with high initial $\bm{\mu}$
increases constraint satisfaction sufficiently, thus mitigating the need for multiple
AL iterations. In summary, we propose the following modifications:

(i) Barrier weights $\mu_i$ are initialized with large values and kept constant.
This reduces eventual constraint satisfaction but avoids numerical ill-conditioning
for very high barrier weights. This also allows using a standard Riccati backward pass
and avoids the higher implementation complexity and runtime of the sqrt-LQR
backward pass (cf.~\cite{Howell2019}). 

(ii) Lagrange-multipliers $\lambda_{k,i}$ are limited by $\lambda_{k,i} \leq
\lambda_{\textrm{max},i}$, thus likewise avoiding numerical issues. In many cases,
this even allows infeasible initializations to be treated robustly.
The violation of constraints resulting from the limited $\bm{\lambda}$ is,
for our application, preferred over obtaining no solution at all.

(iii) We do not apply solution polishing like in \cite{Howell2019}, as it trades increased
runtime for lower constraint violation. However, in our case, faster updates are preferred.

(iv) Except for the first optimization step, we reinitialize AL-ILQR in
MPC fashion with its results from the previous step shifted by the amount
of distance the vehicle has passed. Importantly, this also includes $\bm{\lambda}$,
which allows the Lagrange multipliers to be preserved between planning steps.
This avoids fully recalculating $\bm{\lambda}$ in each planning step, hence 
mitigating the need for multiple AL iterations.

\section{Results}

To investigate the effectiveness of the proposed approach, we subject it
to a simulated as well as a real-world scenario, which showcase
typical situations in urban environments.

\label{sec:evaluation}

\begin{table}[!]
\begin{center}
\caption{Objective parameters used in evaluation}
\label{tbl:parameters}
\begin{tabular}{c|c}
    \hline
    Parameter(s) & Value(s) \\
    \hline
    $v_\textrm{min}$ & \SI{1.0}{m/s} \\
    $a_\textrm{min}, a_\textrm{max}, \hat{a}_\textrm{lat}$ & \SI{\pm 2.5}{m/s^2}, \SI{2.5}{m/s^2} \\
    $j_\textrm{min}, j_\textrm{max}$ & \SI{\pm 1.5}{m/s^3} \\
    $\kappa_\textrm{min}, \kappa_\textrm{max}$ & \SI{\pm 3.0}{m^{-1}} \\
    $w_d$, $w_\kappa$, $w_v$, $w_a$  & 1.0, 20.0, 0.1, 1.0 \\
    $\alpha, \beta$ & $10.0$, $5\cdot10^{-3}$ \\
    \hline
\end{tabular}
\end{center}
\end{table}

\subsection{Implementation}

\begin{algorithm}[!t]
\caption{Real-time Spatial Trajectory Planning}
\label{alg:rstp}
\begin{algorithmic}[1]
\State given ego state, reference line $\bm{\rho}$, environment state
\State $\bm{\rho}^* \leftarrow$ minimize objective \eqref{eq:ref_line_objective} using Alg.~\autoref{alg:solution-strategy} 
\State extract obstacles concerning $\bm{\rho}^*$ 
\State convert $\bm{\rho}^*, \kappa^*$, obstacles into constraints $v_\textrm{lim}$, $t_\textrm{min}$, $t_\textrm{max}$
\State generate $v_\textrm{ref}$ (\autoref{sec:velocity-objective})
\State $v^* \leftarrow$ minimize objective \eqref{eq:profile_objective} using Alg.~\autoref{alg:solution-strategy}
\State execute $(\bm{\rho}^*, v^*)$ via low-level controller
\end{algorithmic}
\end{algorithm}

The proposed objectives and the solution strategy were implemented
in self-contained C code as Python extensions. Derivatives were obtained with
symbolic differentiation via the SymPy package. Further processing steps were
implemented in Python. We assume usage of a perception stack, which generates
the required information for the trajectory planning algorithm. In both scenarios,
the reference line $\bm{\rho}(s)$ is provided from noisy GPS data recorded
on real roads, thus requiring the path optimization from \autoref{sec:path-opt}. 
All steps of the implementation are summarized in Alg.~\autoref{alg:rstp}, where
the low-level controller is executed with a cycle time of \SI{10}{ms}.
If the runtime of the planning step exceeds \SI{10}{ms}, the previous solution is
executed by the controller.
Necessarily, discretization of the integration in the objectives
is required. For all scenarios, we use a discrete step size of $\Delta s =
\SI{0.5}{m}$ and chose $S = \SI{125}{m}$, which results in 250 spatial steps
per objective. Both objective minimizations are terminated if the relative cost change
is $< 10^{-6}$ or after $N = 5$ ILQR iterations to preserve real-time feasibility.
The barrier weights and multiplier limits $\mu = 10^3, \lambda_\textrm{max} = 10^3$ are
used for $t_\textrm{max}$ constraints and $\mu = 10^2, \lambda_\textrm{max} = 10^2$ for
all other constraints. Further parameters were experimentally determined for
comfortable behavior (\autoref{tbl:parameters}).

\subsection{Simulated Scenario}

\begin{figure}[!t]
\centering
\begin{tikzpicture}
\begin{axis}[
    scale only axis,
    width=7.0cm,
    height=2.1cm,
    xlabel={$s$ in \SI{}{m}},
    ylabel={$v$ in \SI{}{m/s}},
    xlabel near ticks,
    ylabel near ticks,
    xmin=0, xmax=125,
    ymin=-1, ymax=12.0,
    legend pos=south west,
    legend columns=-1,
    xmajorgrids=true,
    ymajorgrids=true,
    grid style=solid,
    rounded corners = 1pt,
]

\addplot[
    color=blue!50,
    line width=1.0pt
    ]
    table[x=space,y=lim,col sep=comma] {data/sim_vp_no_smooth.csv};
\addlegendentry{$v_{\textrm{lim}, \kappa}$} 

\addplot[
    color=blue,
    line width=1.0pt
    ]
    table[x=space,y=ref,col sep=comma] {data/sim_vp_no_smooth.csv};
\addlegendentry{$v_{\textrm{ref}, \kappa}$} 
    
\addplot[
    color=red,
    line width=1.5pt
    ]
    table[x=space,y=ref,col sep=comma] {data/sim_vp_smooth.csv};
\addlegendentry{$v_{\textrm{ref}, \kappa^*}$} 

\end{axis}
\end{tikzpicture}
\vspace{-0.7cm}
\caption{Effects of the path objective \eqref{eq:ref_line_objective}: Using local curvature estimates $\kappa$
leads to noisy $v_\textrm{lim}$ and unnecessarily jerky $v_\textrm{ref}$, while $\kappa^*$ yields a smooth reference
profile.
}
\label{fig:sim-path-opt}
\end{figure}
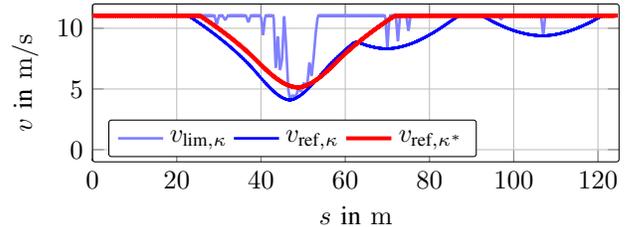
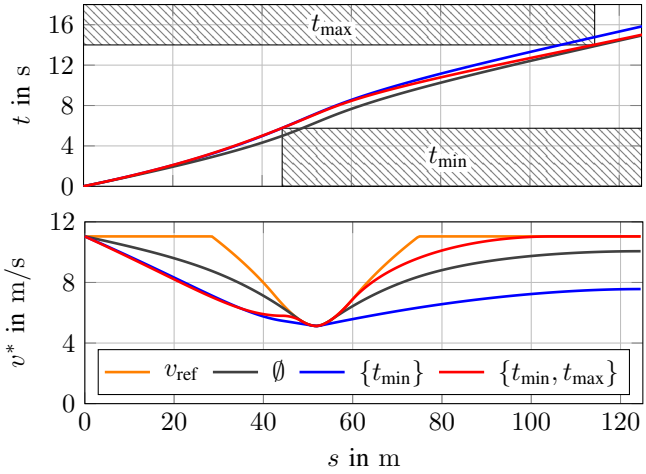
\begin{figure}[!t]
\centering
\subfloat{\begin{tikzpicture}
\begin{axis}[
    width=9cm,
    height=4.0cm,
    ylabel={$t$ in \SI{}{s}},    
    ylabel near ticks,
    xmin=0, xmax=125,
    ymin=0, ymax=18,
    xticklabels={},
    ytick={0,4,8,12,16},
    xmajorgrids=true,
    ymajorgrids=true,
    grid style=solid,
    legend pos=south east,
    enlargelimits=false,
]
\draw[pattern=north west lines, pattern color=gray] (44.5,0) rectangle (125.0,5.75); 
\draw[pattern=north west lines, pattern color=gray] (0,14.0) rectangle (114.5,18.0); 
\node[rectangle, anchor=center, inner sep=1pt, outer sep=0pt] at (55.75, 16.0) {\contour{white}{$t_\textrm{max}$}};
\node[rectangle, anchor=center, inner sep=1pt, outer sep=0pt] at (81.75, 2.875) {\contour{white}{$t_\textrm{min}$}};
\addplot[
    color=blue,
    line width=1.0pt
    ]
    table[x=s,y=t,col sep=comma] {data/sim_veh_path_tmin_constr.csv};
\addplot[
    color=darkgray,
    line width=1.0pt
    ]
    table[x=s,y=t,col sep=comma] {data/sim_veh_path_no_constr.csv};
\addplot[
    color=red,
    line width=1.0pt
    ]
    table[x=s,y=t,col sep=comma] {data/sim_veh_path.csv};
\end{axis}
\end{tikzpicture}}
\vspace{-0.35cm}
\subfloat{\begin{tikzpicture}
\begin{axis}[
    width=9cm,
    height=4cm,
    xlabel={$s$ in \SI{}{m}},
    ylabel={$v^*$ in \SI{}{m/s}},    
    ylabel near ticks,
    xmin=0, xmax=125,
    ymin=0, ymax=12,
    xtick={0,20,40,60,80,100,120},
    ytick={0,4,8,12,16},
    xmajorgrids=true,
    ymajorgrids=true,
    grid style=solid,
    legend columns=-1,
    legend style={at={(0.5, 0.3)},
        anchor=north,
        inner sep=2pt,
        style={column sep=0.1cm}
    }
]

\addplot[
    color=orange,
    line width=1.0pt
    ]
    table[x=space,y=ref,col sep=comma] {data/sim_vp_no_constr.csv};
\addlegendentry{$v_\textrm{ref}$} 

\addplot[
    color=darkgray,
    line width=1.0pt
    ]
    table[x=space,y=opt,col sep=comma] {data/sim_vp_no_constr.csv};
\addlegendentry{$\emptyset$} 

\addplot[
    color=blue,
    line width=1.0pt
    ]
    table[x=space,y=opt,col sep=comma] {data/sim_vp_tmin_constr.csv};
\addlegendentry{\{$t_\textrm{min}$\}} 

\addplot[
    color=red,
    line width=1.0pt
    ]
    table[x=space,y=opt,col sep=comma] {data/sim_vp_all_constr.csv};
\addlegendentry{\{$t_\textrm{min}, t_\textrm{max}$\}} 
    
\end{axis}
\end{tikzpicture}}
\vspace{-0.75cm}
\caption{
    Resulting velocity profiles at $t=0$ and driven $s$-$t$ trajectories for the simulated
    scenario from \autoref{fig:sim-scenario}, with different spatiotemporal constraints applied (none, only $t_\textrm{min}$, $t_\textrm{min}$ and $t_\textrm{max}$). 
    Spatiotemporal constraints are depicted as hatched rectangles.
}
\label{fig:vel-prof-sim}
\end{figure}
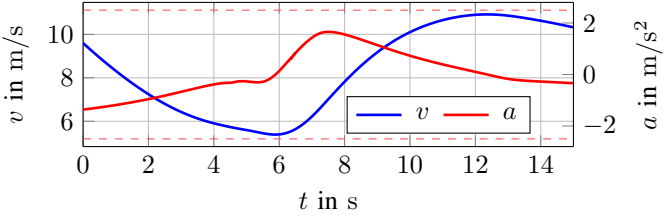
\begin{figure}[!t]
\centering
\begin{tikzpicture}
\begin{axis}[
    width=8.1cm,
    height=3.5cm,
    axis y line*=left,
    ylabel near ticks,
    xmin=0, xmax=15,
    xlabel={$t$ in \SI{}{s}},
    ylabel={$v$ in \SI{}{m/s}},
    xmajorgrids=true,
    ymajorgrids=true,
    grid style=solid,
    xlabel near ticks,
]
\addplot[
    color=blue,
    line width=1.0pt
    ]
    table[x=t,y=v,col sep=comma] {data/sim_veh_vel.csv};
\label{plot-v}
\end{axis}
\begin{axis}[
    width=8.1cm,
    height=3.5cm,
    axis y line*=right,
    axis x line=none,
    xmin=0, xmax=15,
    ymin=-2.8, ymax=2.8,
    ylabel={$a$ in $\SI{}{m/s^2}$},    
    ylabel near ticks,
    yticklabel pos=right,
    legend columns=-1,
    legend style={at={(0.725, 0.35)},
        anchor=north,
        style={column sep=0.10cm}
    }
]
\addlegendimage{/pgfplots/refstyle=plot-v}\addlegendentry{$v$}
\addplot[
    color=red,
    line width=1.0pt
    ]
    table[x=t,y=a,col sep=comma] {data/sim_veh_acc.csv};
\addlegendentry{$a$} 

\addplot[gray, dashed, color=red!70] coordinates {(0,2.5)(15,2.5)};
\addplot[gray, dashed, color=red!70] coordinates {(0,-2.5)(15,-2.5)};

\end{axis}

\end{tikzpicture}
\vspace{-0.7cm}
\caption{
    Velocity and acceleration (with $a_\textrm{min}, a_\textrm{max}$ as dashed lines) of the ego vehicle during the simulated scenario from
    \autoref{fig:sim-scenario}.
}
\label{fig:vel-acc-sim}
\end{figure}

The simulated experiments were executed on an AMD 1900X CPU @ \SI{3.8}{GHz} base clock using a
single thread. Perfect knowledge of the states of all vehicles is assumed. The vehicle movement is
simulated with a kinematic bicycle model. The complex scenario from \autoref{fig:sim-scenario} is examined here.
Additional scenarios can be found on the project website \href{https://github.com/joruof/tpl}{https://github.com/joruof/tpl}. 

\longautoref{fig:sim-scenario} depicts the scenario setup at $t=0$.
The ego vehicle has to merge smoothly onto a main road without stopping while providing
right of way to another traffic participant traveling at the speed limit $v_\textrm{lg} =\SI{40}{km/h}$.
As an additional complication, a traffic light has to be passed after the intersection,
which is scheduled to turn red at $t=\SI{14}{s}$. \longautoref{fig:sim-path-opt} shows that
the path optimization from \autoref{sec:path-opt} leads to accurate curvature
estimation and enables the generation of a smooth $v_\textrm{ref}$ profile.
Given the optimized path, two spatiotemporal constraints 
$t_\textrm{min}(\SI{44.5}{m}) = \SI{5.75}{s}$, $t_\textrm{max}(\SI{114.5}{m}) = \SI{14.0}{s}$
are generated. \longautoref{fig:vel-prof-sim} compares the resulting velocity profiles for different
spatiotemporal constraint sets. Given no spatiotemporal constraints ($\emptyset$), the intersection is reached too early,
resulting in a collision. For $\{t_\textrm{min}\}$ only, the intersection is entered safely, but the 
green traffic light is missed. With $\{t_\textrm{min}, t_\textrm{max}\}$ applied, a solution
is found, which first reduces the velocity at $\SI{44.5}{m}$ to keep the $t_\textrm{min}$ constraint, 
then closely follows the reference velocity $v_\textrm{ref}$ to reach the traffic light in time.
As \autoref{fig:vel-prof-sim} and \autoref{fig:vel-acc-sim} show, all acceleration,
velocity, and spatiotemporal constraints are satisfied by the solution. Runtime statistics
for our approach in the simulated scenario are given in \autoref{tbl:simulated-runtime}.
The maximum total runtime for the proposed method, which includes preprocessing, is below \SI{10}{ms}.
The results show that with the proposed method even multiple conflicting constraints can be handled
robustly and  efficiently.

\begin{table}[!t]
\vspace{-0.3cm}
\centering
\caption{Runtimes of our method in the simulated scenario from \autoref{fig:sim-scenario}.}
\label{tbl:simulated-runtime}
\begin{tabular}{c|c|c|c}
    \hline
    Component & Mean & Stddev. & Max. \\
    \hline 
    lateral (\autoref{eq:ref_line_objective}) 
       & \SI{0.354}{ms} & \SI{0.116}{ms} & \SI{1.50}{ms} \\
    longitudinal (\autoref{eq:profile_objective}) 
       & \SI{0.214}{ms} & \SI{0.101}{ms} & \SI{1.380}{ms} \\
    total (lat. + lon. + preproc.) &  \SI{1.4}{ms} & \SI{0.290}{ms} & \SI{9.610}{ms} \\
    \hline
\end{tabular}
\end{table}

\subsection{Real World Scenario}

\begin{figure}[!t]
\centering
\pgfplotsset{
compat=newest,
}
\begin{tikzpicture}
\begin{axis}[
    title={},
    width=9cm,
    height=3cm,
    xlabel={$t$ in \SI{}{s}},
    ylabel={$v$ in \SI{}{m/s}},
    ylabel near ticks,
    xmin=0, xmax=41,
    ymin=-1, ymax=12,
    ytick={0,4,8,12,16},
    legend pos=south east,
    xmajorgrids=true,
    ymajorgrids=true,
    grid style=solid,
]
\addplot[
    color=blue,
    line width=1.0pt
    ]
    table[x=t,y=v,col sep=comma] {data/real/real_velocity_offset.csv};

\addplot[black] coordinates {(0,-1)(0,12)};
\addplot[black] coordinates {(5.905,-1)(5.905,12)};
\addplot[black] coordinates {(16.681,-1)(16.681,12)};
\addplot[black] coordinates {(25.318,-1)(25.318,12)};
\addplot[black] coordinates {(37.715,-1)(37.715,12)};

\node[circle, draw=black, fill=white, anchor=west, inner sep=1pt, outer sep=3pt] at (0,6) {1};
\node[circle, draw=black, fill=white, anchor=west, inner sep=1pt, outer sep=3pt] at (5.905,6) {2};
\node[circle, draw=black, fill=white, anchor=east, inner sep=1pt, outer sep=3pt] at (16.681,6) {3};
\node[circle, draw=black, fill=white, anchor=west, inner sep=1pt, outer sep=3pt] at (25.318,3) {4};
\node[circle, draw=black, fill=white, anchor=east, inner sep=1pt, outer sep=3pt] at (37.715,3) {5};
\end{axis}
\end{tikzpicture}
\vspace{-0.5cm}
\caption{
    Vehicle velocity during the real-world scenario.
    Events 1-5 are marked by numbers and detailed in \autoref{fig:real-vps}. 
}
\label{fig:real-vel}
\end{figure}
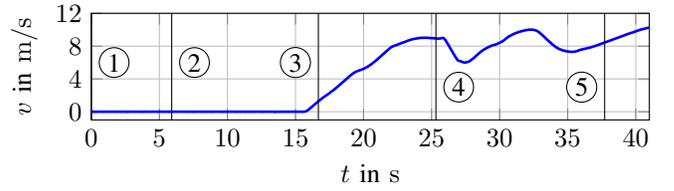

\begin{figure}[!t]
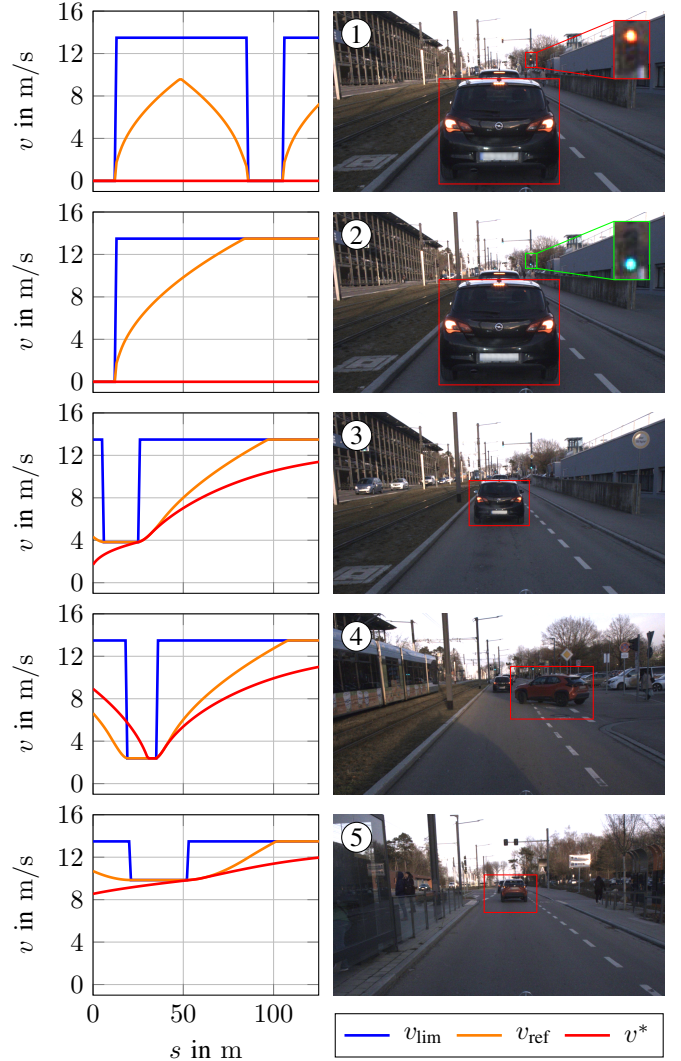

\centering
\subfloat{%
\input{plots/real/real_vp_start}
}
\vspace{-0.55cm}

\subfloat{%
\input{plots/real/real_vp_green}
}
\vspace{-0.55cm}

\subfloat{%
\input{plots/real/real_vp_follow}
}
\vspace{-0.55cm}

\subfloat{%
\input{plots/real/real_vp_cut_in}
}
\vspace{-0.55cm}

\subfloat{%
\input{plots/real/real_vp_follow_2}
}
\vspace{-0.6cm}
\caption{
    The left column shows the results of the velocity profile optimization for events 1-5 
    (\autoref{fig:real-vel}) during the real-world scenario, where $s=0$ is the respective current 
    position of the ego vehicle. For reference, the right column depicts images from the front camera
    of the ego vehicle, where relevant objects are marked by boxes.
}
\label{fig:real-vps}
\end{figure}

The real-world scenario was performed with an automated test vehicle of our institute,
a Mercedes-Benz S-Class equipped with additional sensors (e.g. cameras, LiDARs).
Steering and brake actuators as well as the drive train are controlled by our planning approach.
The vehicle state was obtained by a DGPS unit with centimeter accuracy.
All computations are performed by an onboard PC with an AMD 3990X CPU @ \SI{2.9}{GHz} base clock,
using a single thread for our planner.

As real-world scenario, planning in a dense urban environment is demonstrated.
The ego vehicle has to follow a reference line while reacting to the environment and
other traffic participants. \longautoref{fig:real-vel} shows the recorded vehicle
velocity, where 5 events are marked. The corresponding velocity profiles for the events
of the vehicle are shown together with front camera images in \autoref{fig:real-vps}.

\emph{Event 1:} Initially, the ego vehicle is stopped behind another car, waiting in front
of a red light. The resulting velocity profile shows two sections with $0$ velocity,
corresponding to the leading vehicle and the traffic light.

\emph{Event 2:} At $t=\SI{5.9}{s}$, the traffic light turns green, which removes the
second section of $v=0$. As the leading vehicle has not started moving yet,
the velocity is still $0$ at the beginning of the velocity profile.

\emph{Event 3:} As the leading vehicle starts moving, the $v \leq v_\textrm{ref}$ constraint 
is successively lifted, which allows the optimization to increase the velocity,
yielding smooth following behavior.

\emph{Event 4:} A car unexpectedly cuts in front of the ego vehicle, which
makes the solution fall below the safety distance thus violating the velocity
constraints. In such situations, barrier function
constraints \cite{Chen2018A} quickly raise costs to very large (or undefined) values making the
optimization numerically unstable \cite{Chen2019}. Due to the proposed limitations of 
$\bm{\lambda}, \bm{\mu}$, our adapted solution strategy (cf.~\autoref{sec:solution_strategy})
remains numerically stable and is able to react adequately. The solution $v^*$ applies 
minimum acceleration until the constraint is fulfilled while keeping the remainder of the 
trajectory consistent. 

\emph{Event 5:} After the cut-in, the ego vehicle continues following the leading
vehicle normally until the end of the scenario.

\autoref{tbl:real-runtime} lists the runtimes of our method recorded during the real-world scenario.
Due to a higher number of objects that need to be preprocessed, the total computation time is higher
compared to the simulated scenario, exceeding \SI{10}{ms} at the maximum. Still, $97.4 \%$
of planning steps take less than \SI{10}{ms}, showing that our approach can rapidly provide trajectory
updates, even in complicated real-world scenarios.
\begin{table}[!t]
\centering
\caption{Runtimes of the planer in the real-world scenario.}
\label{tbl:real-runtime}
\begin{tabular}{c|c|c|c}
    \hline
    Component & Mean & Stddev. & Max. \\
    \hline 
    lateral (\autoref{eq:ref_line_objective}) 
       & \SI{0.529}{ms} & \SI{0.195}{ms} & \SI{1.371}{ms} \\
    longitudinal (\autoref{eq:profile_objective}) 
       & \SI{0.924}{ms} & \SI{0.335}{ms} & \SI{2.371}{ms} \\
    total (lat. + lon. + preproc.) &  \SI{6.796}{ms} & \SI{1.627}{ms} & \SI{14.15}{ms} \\
    \hline
\end{tabular}
\end{table}

\section{Conclusion}
\label{sec:conclusion}

We have presented a dynamic optimization approach for foresighted trajectory
planning, which can handle general velocity as well as spatiotemporal constraints. 
By choosing a spatial formulation, a fast, tailored AL-ILQR method could be applied,
resulting in small cycle times and thus substantially reduced reaction time in
complex urban environments. Since the proposed formulation currently plans longitudinally
along a provided reference line, future work will focus on extensions that allow for
lateral deviations (e.g. for overtaking) and further reduction of the maximum computation time.

\bibliography{root.bib}
\bibliographystyle{ieeetr}

\end{document}